\documentclass[conference]{IEEEtran}
\IEEEoverridecommandlockouts

\usepackage{listings}
\usepackage{graphicx}
\usepackage{subcaption}
\usepackage{cite}
\usepackage{amsmath,amssymb,amsfonts}
\usepackage[font=small]{caption}
\usepackage{algorithmic}
\usepackage{textcomp}
\usepackage{xcolor}
\def\BibTeX{{\rm B\kern-.05em{\sc i\kern-.025em b}\kern-.08em
    T\kern-.1667em\lower.7ex\hbox{E}\kern-.125emX}}

\begin{document}

\title{A Computer Vision Pipeline for Iterative Bullet Hole Tracking in Rifle Zeroing}

\author{\IEEEauthorblockN{Robert M. Belcher, Brendan C. Degryse, Leonard R. Kosta, Christopher J. Lowrance}
\IEEEauthorblockA{\textit{Department of Electrical Engineering and Computer Science} \\
\textit{United States Military Academy}\\
West Point, NY \\
\{robert.belcher, brendan.degryse, leonard.kosta, christopher.lowrance\}@westpoint.edu}
}

\maketitle

\IEEEpubid{%
  \begin{minipage}{\textwidth}\ \\[12pt]
    \makebox[0pt][l]{U.S. Government work not protected by U.S. copyright}
  \end{minipage}
}

\IEEEpubidadjcol

\begin{abstract}
Adjusting rifle sights, a process commonly called ``zeroing,'' requires shooters to identify and differentiate bullet holes from multiple firing iterations. Traditionally, this process demands physical inspection, introducing delays due to range safety protocols and increasing the risk of human error. We present an end-to-end computer vision system for automated bullet hole detection and iteration-based tracking directly from images taken at the firing line. Our approach combines YOLOv8 for accurate small-object detection with Intersection over Union (IoU) analysis to differentiate bullet holes across sequential images. To address the scarcity of labeled sequential data, we propose a novel data augmentation technique that removes rather than adds objects to simulate realistic firing sequences. Additionally, we introduce a preprocessing pipeline that standardizes target orientation using ORB-based perspective correction, improving model accuracy. Our system achieves 97.0\% mean average precision on bullet hole detection and 88.8\% accuracy in assigning bullet holes to the correct firing iteration. While designed for rifle zeroing, this framework offers broader applicability in domains requiring the temporal differentiation of visually similar objects.
\end{abstract}


\section{Introduction}

Rifle marksmanship requires precise sight alignment, typically achieved through a process known as zeroing: iteratively adjusting a rifle's sights to ensure that bullets strike the intended point of aim. Zeroing involves firing a series of rounds, inspecting the resulting bullet holes on the target, and making sight corrections based on shot placement. This process must be repeated any time a rifle is reassembled, transported, or assigned to a new shooter.

This routine is complicated by the need for strict safety protocols on firing ranges. To inspect their targets, shooters must cease fire and walk downrange, which introduces a form of mutual exclusion: either all shooters are firing, or all are inspecting. This constraint mirrors the classic readers-writers problem in concurrent systems: shooters act as ``writers'' while firing and ``readers'' while inspecting, but cannot do both concurrently. These enforced transitions between reading and writing phases slow down the entire range operation and introduce opportunities for human error in bullet hole identification.

An automated system that detects and differentiates bullet holes directly from images taken at the firing line could eliminate this bottleneck. Such a system must answer two key questions: (1) Where are the bullet holes? and (2) Which holes belong only to the most recent iteration? Existing object detection methods, including classical image processing and modern deep learning approaches, can answer the first question but not the second. Differentiating visually similar bullet holes across firing iterations remains a challenge.

\begin{figure}[h]
\centering
\includegraphics[width=\linewidth]{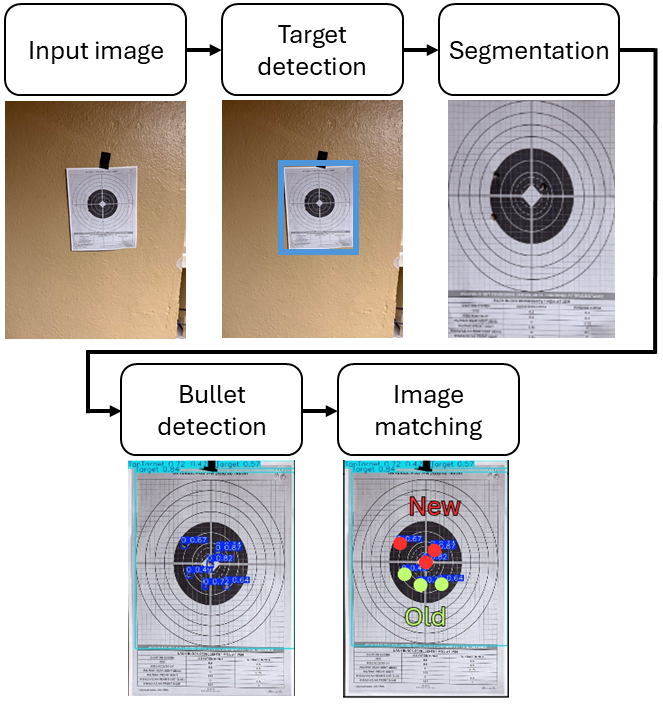}
\caption{Overview of the bullet hole tracking pipeline.}
\label{fig:pipeline}
\end{figure}

In this work, we present an end-to-end computer vision pipeline for automated bullet hole detection and per-iteration tracking. Our system uses You Only Look Once (YOLOv8)~\cite{10533619} for bullet hole detection and introduces a novel application of Intersection over Union (IoU) to distinguish new bullet holes from those in prior images. To support this, we develop a preprocessing pipeline using Oriented FAST and Rotated BRIEF (ORB)~\cite{rublee2011orb} for target segmentation and apply a novel data augmentation strategy that removes rather than adds objects to simulate realistic firing sequences. An overview of the full pipeline, from raw image to iteration-labeled bullet holes, is shown in Fig.~\ref{fig:pipeline}. Our results demonstrate competitive detection accuracy and robust iteration tracking.

This paper makes the following contributions:

\begin{itemize}
    \item We introduce a novel image processing pipeline that detects and tracks bullet holes across firing iterations using YOLOv8 and IoU analysis.
    \item We propose a unique data augmentation technique that enables synthetic generation of firing sequences by removing bullet holes.
    \item We demonstrate a preprocessing approach using ORB and perspective transforms to normalize images across camera angles and improve detection performance.
    \item We report competitive detection and tracking accuracy, achieving 97.0\% mAP50 and 88.8\% firing iteration classification accuracy on our dataset.
\end{itemize}

The rest of the paper is structured as follows: Section II reviews related work and foundational concepts in bullet hole detection, object tracking, and IoU. Section III describes our methods, including data collection, augmentation, segmentation, and tracking. Section IV presents experimental results. Section V discusses broader implications and Section VI outlines future work.

\section{Background and Related Work}
The task of bullet hole detection in target images lies at the intersection of object detection, image registration, and object tracking. This section outlines relevant techniques in these areas and highlights the limitations of prior work that motivate our approach.

\subsection{Bullet Hole Detection in Prior Work}
Recent research has applied both classical image processing and deep learning to bullet hole detection. Early methods employed contour analysis, edge detection, and Hough transforms to identify circular features, but struggled under conditions of occlusion, overlap, or inconsistent lighting.

Deep learning models such as Mask R-CNN and Faster R-CNN have since improved detection accuracy in complex scenes. For example, Butt et al. \cite{butt2024application} applied YOLOv8 and Detectron2 to achieve a mAP50 of 96.7\% on bullet hole detection, while Ahmed et al. \cite{10080859} reported 83.3\% average precision using Mask R-CNN. However, most prior work focuses on static detection, identifying bullet holes in a single image. Little work has been done on tracking bullet holes across multiple firing iterations.

A key challenge in iteration tracking is distinguishing new bullet holes from previous ones, especially when shots cluster in tight groups or photos capture the same target from different perspectives. No existing system, to our knowledge, robustly tracks bullet holes across firing rounds using field-acquired imagery.

\subsection{YOLO for Object Detection}

YOLO is a real-time object detection architecture that partitions an image into a grid and predicts bounding boxes and class probabilities in a single pass through a convolutional neural network \cite{redmon2016lookonceunifiedrealtime}. YOLO’s speed and accuracy make it well-suited for detecting small objects such as bullet holes, especially in real-world settings with visual clutter. Our work uses YOLOv8, which offers enhanced performance for small-object detection through improved spatial resolution and architectural refinements \cite{10533619}.

\subsection{Tracking Across Images with IoU}

Intersection over Union (IoU) is a standard metric for evaluating object detector performance. It measures the overlap between a predicted bounding box and the ground truth:

\begin{equation}
IoU = \frac{\text{Area of Overlap}}{\text{Area of Union}}
\end{equation}

Although IoU is typically used as a threshold during evaluation (e.g., mAP50 denotes predictions with $\ge$50\% IoU), we adapt IoU as a matching function to associate bullet holes across sequential images. This enables us to infer which holes are new in a given iteration by comparing detections frame to frame. This application of IoU differs from its traditional use and allows us to track persistent objects even without fixed camera positioning, meaning that each sequential image could be taken from a mobile device.

\subsection{Target Segmentation via ORB}

To ensure accurate spatial alignment across images taken from different angles or distances, we normalize each image using perspective correction. We use ORB, a fast feature detector and descriptor, to identify the corners of the target and align them via homography. This ensures consistent localization of bullet holes across firing iterations, even when the camera is handheld or repositioned.

\section{Methods}

The system processes raw images taken at the firing line and outputs the locations of bullet holes, labeled by firing iteration. Our work includes dataset construction, data augmentation and preprocessing, bullet hole detection, and inter-image tracking. This section describes each component of the system and the evaluation metrics used to assess performance.

\subsection{Dataset Construction}

We constructed a diverse dataset from three sources:

\begin{enumerate}
    \item Operational Range Data: We collaborated with a U.S. Army unit at Fort Benning to collect over 280 images from 25-meter rifle qualification ranges. These images varied in quality, resolution, and lighting, and did not include ground-truth annotations across firing iterations.
    \item External Labeled Data: To supplement our training data, we incorporated 1,243 labeled images from a publicly available dataset by Butt et al. \cite{butt2024application}, which includes bounding boxes for bullet holes but not sequential image pairs.
    \item Synthetic and Controlled Data: To evaluate bullet hole tracking across iterations, we generated synthetic sequences by simulating firing events in two stages. First, we created a pencil-based synthetic dataset. We used a No. 2 pencil (7mm diameter) to create 3–4 holes in paper targets, captured an image, added 3–4 more holes, and captured a second image. Second, we conducted a live-fire collection using an AR-15 chambered in 5.56×45mm. For each target, we fired two sets of 3 to 4 rounds, capturing an image after each. This yielded 11 additional image pairs under realistic ballistic and lighting conditions. In total, we constructed a dataset of 33 labeled two-iteration sequences (22 pencil-hole targets, 11 live-fire targets), used for evaluating bullet-hole tracking across iterative firing rounds.
\end{enumerate}

\subsection{Data Augmentation and Preprocessing}

To address the scarcity of labeled, sequential bullet hole data, we explored multiple augmentation and preprocessing strategies designed to simulate realistic firing sequences and normalize image conditions.

Early experiments attempted to synthetically add bullet holes to images using generative AI tools, but the results were visually unrealistic. We instead developed a reverse augmentation strategy: selectively removing bullet holes from real images using AI-powered image editing. By creating a ``clean'' version of a target and comparing it to the original, we could simulate the appearance of new bullet holes in a second iteration. This method produced visually plausible iteration pairs, but both images shared an identical camera perspective, limiting viewpoint diversity.

To ensure robustness across more realistic conditions, such as images taken from different distances or angles, we implemented a perspective normalization pipeline. For each image, we used ORB to detect the corners of the target and computed a homography to transform the target into a standardized reference frame. This allowed consistent spatial alignment of bullet holes across iterations, even when the images were captured from different viewpoints.

If ORB fails to segment the target, typically due to occlusion or visual noise, a second-pass fallback using YOLOv8 provides a rough bounding box for reattempting ORB-based perspective correction. This two-step segmentation approach improved robustness without requiring explicit manual cropping.

\subsection{Bullet Hole Detection Using YOLOv8}

We used YOLOv8 for bullet hole detection due to its strong performance on small-object localization tasks. We fine-tuned the model on a combination of normalized images from our custom datasets and unaltered images from the external dataset provided by Butt et al. \cite{butt2024application}.

Training was performed over 100 epochs using default hyperparameters. The model was applied to segmented target regions to generate bounding boxes around bullet holes. This same YOLOv8 model was also used as a fallback to detect targets during preprocessing when ORB segmentation failed.

By restricting bullet hole detection to normalized target regions, we improved accuracy compared to applying YOLO across the full, unprocessed image.

\subsection{Tracking Bullet Holes Across Firing Sequences}

To distinguish new bullet holes from those present in previous iterations, we used Intersection over Union (IoU) as a matching function. For each bounding box detected in the second image of a firing sequence, we computed its IoU with all bounding boxes from the first image. If the maximum IoU exceeded a threshold of 0.5, the bullet hole was considered a match with a previous shot; otherwise, it was classified as new.

This method relies on the assumption that, after perspective normalization, bullet holes will appear in approximately the same location across iterations. This is a reasonable assumption because the paper target remains in fixed position between iterations, and our normalization process aligns all images to a common reference frame using the known geometry of the target. As a result, any visual displacement of bullet holes between iterations is minimal. Under these conditions, IoU provides a simple and effective method for matching bullet holes across sequential images without requiring more complex techniques such as feature descriptors or optical flow. This approach allowed us to isolate new bullet holes introduced during each firing iteration and laid the groundwork for computing group centers, shot spread, and suggested sight adjustments.

\subsection{Evaluation Metrics}

We evaluated the performance of the system at both component and pipeline levels using multiple standard metrics.

For detection quality, we report mean average precision at IoU thresholds of 0.5 (mAP50) and across the full range from 0.5 to 0.95 (mAP50–95), for both bullet hole and target detection.
For tracking, we use the Jaccard Index to measure the overlap between predicted and ground-truth sets of new bullet holes. Firing iteration classification accuracy is computed as the percentage of bullet holes correctly classified as belonging to the current firing iteration.

Finally, we define full pipeline accuracy as the proportion of test cases in which the system correctly detects the target, identifies the correct number of bullet holes, and classifies each one to the appropriate iteration. These metrics are discussed in detail in the next section.

\section{Results}

We evaluated our system across four levels of performance: (1) segmentation accuracy, (2) bullet hole and target detection, (3) tracking across firing iterations, and (4) end-to-end pipeline accuracy. Results are reported on a combination of real and synthetic datasets described in Section III.

\subsection{Segmentation Accuracy}

We first assessed the ability of the preprocessing pipeline to correctly segment targets from the input images. On images where the entire target was fully visible, the ORB-based perspective correction successfully segmented 320 out of 439 targets, yielding a segmentation accuracy of 72.9\%. Most failures occurred in cases where targets were physically overlapping or occluded, which can be mitigated operationally by ensuring targets do not overlap on the range.

Despite these challenges, the segmentation algorithm handled extreme cases well, including images with up to 12 distinct targets. In nearly all cases where the entire target was visible and unobstructed, segmentation succeeded regardless of scale.

\subsection{Bullet Hole and Target Detection}

We evaluated detection performance using standard mAP metrics. When applied to segmented target regions, our YOLOv8 model achieved a mAP50 of 97.0\% for bullet hole detection. This matches the best reported performance in the literature, 96.7\% \cite{butt2024application}. Our marginal improvement may be attributed to our use of a larger and more diverse dataset; however, we acknowledge that direct comparison is limited by differences in evaluation data. Detection performance remained strong across a range of IoU thresholds, with mAP50–95 reaching 63.3\%.

For target detection (used as a fallback in the segmentation stage), the model achieved a mAP50 of 62.2\% and mAP50–95 of 52.9\%. This lower performance is attributable to the limited number of labeled target examples in our dataset, but it was sufficient for robust fallback behavior during preprocessing.

These results confirm that our model performs at state-of-the-art levels for bullet hole detection, especially when used on segmented, normalized inputs.

\subsection{Firing Iteration Classification Accuracy}

To evaluate tracking between iterations, we applied the IoU-based matching process described in Section III to all sequential image pairs. Bullet holes in the second image of each pair were compared to those in the first image, and assigned to the current iteration if no match exceeded an IoU threshold of 0.5.

\begin{figure}[h]
\centering
\includegraphics[width=\linewidth]{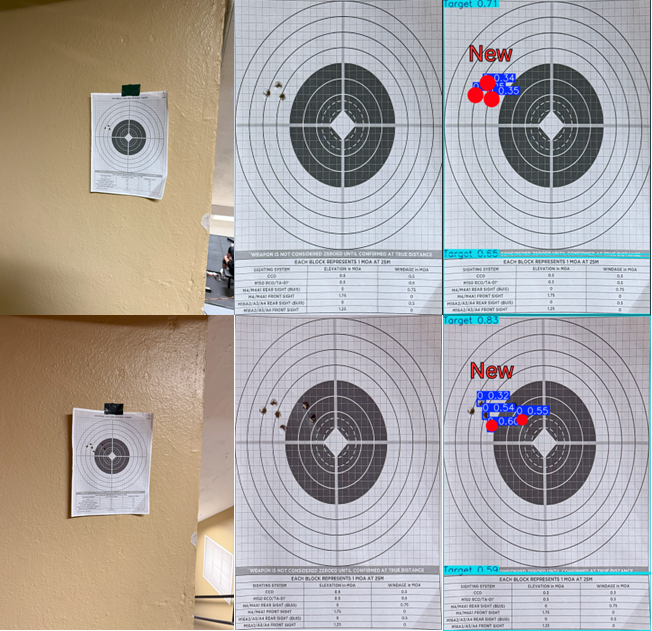}
\caption{Example output over two firing iterations. The top row shows images from the first firing iteration, and the bottom row shows images from the second iteration. These examples were collected indoors as part of a controlled evaluation.}
\label{fig:tracking_example}
\end{figure}

Across all test pairs, the average Jaccard Index between predicted and ground-truth sets of “new” bullet holes was 0.88, indicating high consistency in bullet hole classification between iterations. The system assigned 88.8\% of bullet holes to the appropriate firing iteration. Fig.~\ref{fig:tracking_example} illustrates two examples of this process, showing how the system identifies newly created bullet holes across firing iterations. These results show that the system can reliably distinguish new bullet holes, even across varied viewpoints.

\subsection{End-to-End System Accuracy}

To measure full pipeline performance, we evaluated the system’s ability to take a raw input image, detect targets, segment them correctly, detect all bullet holes, and assign them to the correct iteration. When applied to the full dataset, the system correctly identified both the target and all bullet holes with correct iteration classification in 40.3\% of cases.

This holistic metric reflects the compounding effect of minor errors at each stage of the pipeline. Despite this, the high per-component metrics (e.g., 97.0\% mAP50, 88.8\% tracking accuracy) suggest the system is reliable in operational conditions and provides actionable insights at each iteration.

Table \ref{tab:summary_results} summarizes key performance metrics across segmentation, detection, tracking, and full pipeline evaluation.

\begin{table}[h]
\centering
\caption{Summary of system performance metrics}
\begin{tabular}{|l|c|}
\hline
\textbf{Metric} & \textbf{Value} \\
\hline
Segmentation Accuracy (targets fully in frame) & 72.9\% \\
\hline
Bullet Hole Detection mAP50 & 97.0\% \\
\hline
Bullet Hole Detection mAP50–95 & 63.3\% \\
\hline
Target Detection mAP50 & 62.2\% \\
\hline
Target Detection mAP50–95 & 52.9\% \\
\hline
Average Jaccard Index (tracking) & 0.88 \\
\hline
Firing Iteration Classification Accuracy & 88.8\% \\
\hline
Full Pipeline Accuracy & 40.3\% \\
\hline
\end{tabular}
\label{tab:summary_results}
\end{table}

\section{Discussion}

This work demonstrates a novel, effective solution for detecting and tracking bullet holes across sequential firing iterations. Our results show that each component of the system—segmentation, detection, and tracking—performs robustly under real-world conditions, with bullet hole detection achieving 97.0\% mAP50 and firing iteration classification reaching 88.8\% accuracy. These metrics rival the performance of prior work on bullet hole detection and, to our knowledge, represent the first demonstration of accurate bullet hole tracking across iterations using real or simulated sequential imagery.

A key insight is that detection performance improves significantly when images are preprocessed via target segmentation and perspective normalization. Segmenting the region of interest allowed the YOLOv8 model to focus on small, dense visual features with high accuracy. Even under challenging conditions, such as slight changes in camera angle or image resolution, the preprocessing pipeline enabled consistent spatial alignment across iterations, which proved essential for reliable tracking.

The use of Intersection over Union (IoU) as a matching function, rather than purely as an evaluation metric, proved to be a lightweight yet effective strategy for temporal association. This approach eliminates the need for more complex and error-prone tracking techniques such as optical flow or re-identification models, and it adapts well to the constraints of field environments.

Although the full pipeline accuracy (40.3\%) is lower than the individual component performance, this is expected given the compounding nature of errors across stages. Importantly, in most failure cases, intermediate outputs such as bullet hole detection or iteration classification still provide actionable insight to the user. The system is therefore usable, particularly when deployed as a decision-support tool rather than a fully autonomous scoring system.


\section{Conclusions and Future Work}

We presented a novel computer vision pipeline for detecting and tracking bullet holes across firing iterations, enabling shooters to identify new shot groups from images taken at the firing line. Our system combines YOLOv8 with perspective normalization and a new application of IoU for temporal differentiation, achieving improved detection accuracy and high iteration tracking reliability. While developed for rifle zeroing, the core techniques of data augmentation via object removal and inter-frame tracking using IoU generalize to a range of image comparison tasks. This work lays the foundation for real-time, image-based firearm calibration systems and opens avenues for broader applications in visual inspection, diagnostics, and automated feedback.

Several extensions could improve the system. Although the current pipeline is designed for batch processing, the speed of YOLOv8 opens the door for real-time deployment. Integrating the system into a mobile application or web interface would allow shooters to receive immediate visualizations and sight adjustment recommendations from the firing line, removing the need for downrange inspection. It would also allow them to provide labels for new training data. We also plan to explore deep learning-based super-resolution techniques to improve detection quality on degraded or compressed images.

\section*{Acknowledgment}
The views expressed in this article are those of the authors and do not reflect the official policy or position of the Department of the Army, Department of Defense or the U.S. Government.

\nocite{*}
\bibliographystyle{IEEEtran}
\bibliography{thesis.bib}

\end{document}